\documentclass[10pt,twocolumn,letterpaper]{article}

\usepackage[pagenumbers]{cvpr} 
\usepackage{setspace}
\usepackage[dvipsnames]{xcolor}

\definecolor{cvprblue}{rgb}{0.21,0.49,0.74}
\usepackage[pagebackref,breaklinks,colorlinks,citecolor=cvprblue]{hyperref}

\def\paperID{16762} 
\def\confName{CVPR}
\def\confYear{2024}

\title{Learn and Search: An Elegant Technique for Object Lookup using Contrastive Learning}

\author{Chandan Kumar \textsuperscript{*}\\
Iowa State University\\
Ames, IA 50011, USA\\
{\tt\small chandan@iastate.edu}
\and
Jansel Herrera-Gerena \thanks{ Equal Contribution}\\
Iowa State University\\
Ames, IA 50011, USA\\
{\tt\small janselh@iastate.edu}
\and
John Just\\
Iowa State University\\
Ames, IA 50011, USA\\
{\tt\small justjo@iastate.edu}
\and
Ali Jannesari\\
Iowa State University\\
Ames, IA 50011, USA\\
{\tt\small jannesar@iastate.edu}
\and
Matthew Darr\\
Iowa State University\\
Ames, IA 50011, USA\\
{\tt\small darr@iastate.edu}
}

\begin{document}
\maketitle
\begin{abstract}

The rapid proliferation of digital content and the ever-growing need for precise object recognition and segmentation have driven the advancement of cutting-edge techniques in the field of object classification and segmentation. This paper introduces "Learn and Search", a novel approach for object lookup that leverages the power of contrastive learning to enhance the efficiency and effectiveness of retrieval systems.

In this study, we present an elegant and innovative methodology that integrates deep learning principles and contrastive learning to tackle the challenges of object search. Our extensive experimentation reveals compelling results, with "Learn and Search" achieving superior Similarity Grid Accuracy, showcasing its efficacy in discerning regions of utmost similarity within an image relative to a cropped image.

The seamless fusion of deep learning and contrastive learning to address the intricacies of object identification not only promises transformative applications in image recognition, recommendation systems, and content tagging but also revolutionizes content-based search and retrieval. The amalgamation of these techniques, as exemplified by "Learn and Search," represents a significant stride in the ongoing evolution of methodologies in the dynamic realm of object classification and segmentation.



\end{abstract}    
\section{Introduction}
\label{sec:intro}



In the ever-evolving realm of technology and the widespread integration of digital devices, the surge in digital content creation has reached unprecedented heights. As this reservoir of digital assets expands at a rapid pace, the imperative to efficiently search and retrieve relevant images becomes increasingly pronounced. At the heart of this pursuit lies the fundamental challenge of image retrieval, where users endeavor to locate images aligned with specific conceptualizations.

The effective communication of these conceptualizations to retrieval systems stands as a central predicament in image retrieval. Various methods, ranging from textual descriptions to visually analogous images, sketches, or a combination thereof, are employed to articulate these concepts in the form of search queries. Image retrieval, often referred to as the identification of images comparable to a given query image, emerges as a cornerstone in the realm of computer vision.

Deep learning has made significant strides in image retrieval across diverse applications, spanning cloth retrieval \cite{10.1145/2671188.2749318}, biomedical image retrieval \cite{Dubey2019LocalBD}, face retrieval \cite{DONG2018357} \cite{Dubey_2021}, remote sensing image retrieval \cite{9069909}, landmark retrieval \cite{9022111}, social image retrieval \cite{9115089}, and video retrieval \cite{Zhang2016PlayAR}. However, amidst these technological advancements, the indispensability of humans in the image-retrieval system remains a constant. An overarching goal of this endeavor is to address the time-related burdens associated with human annotation labor, prompting the exploration of object lookup through fully unsupervised learning methodologies.

Beyond technological advancements, our work resonates with high social impacts by endeavoring to alleviate the time-intensive nature of human annotation labor. The prospect of saving valuable human resources, coupled with the transformative potential of our fully unsupervised learning approach, positions our work at the intersection of cutting-edge technology and societal welfare. As we navigate through the intricate landscape of image retrieval challenges, our innovative methodology, "Learn and Search," emerges as a beacon of progress, not only redefining object retrieval systems but also contributing significantly to the conservation of human annotation labor.
\section{Related Works}
\label{related_works}
\textbf{Unsupervised Image Retrieval:} Traditional approaches to unsupervised image retrieval conventionally adhere to a structured two-step methodology. In the initial phase, paramount importance is given to feature extraction, where intricate features are meticulously derived from input images using handcrafted descriptors such as GIST \cite{Oliva2001ModelingTS} and SIFT \cite{Lowe2004DistinctiveIF}. Following this, the matching stage employs advanced techniques, including binary hashing [\cite{10.1145/509907.509965}, \cite{SALAKHUTDINOV2009969}, \cite{NIPS2008_d58072be}] or product quantization [\cite{6909519}, \cite{6619223}, \cite{5432202}], to effectuate a transformation of the embedding space. This transformation, whether into Hamming space or a Cartesian product of subspaces, facilitates efficient image retrieval. Notwithstanding, recent advancements in deep learning, such as deep hashing \cite{li2017deep}, \cite{yuan2020central}, and deep product quantization \cite{klein2020endtoend}, \cite{10.1007/s11263-020-01326-x}, have asserted dominance, outperforming their traditional counterparts.

Despite the enhanced performance of these methods, their efficacy is contingent on the availability of labeled data for model training. Addressing this limitation, several deep unsupervised learning models \cite{jang2022selfsupervised}, \cite{wang2022contrastive} have emerged. While these models exhibit promise in an unsupervised setting, their dependence on a pre-trained feature encoder remains a notable characteristic.

Our research is situated within the domain of deep unsupervised learning, introducing a comprehensive end-to-end methodology that is entirely reliant on unsupervised learning principles. Our methodology draws inspiration from Contrastive learning \cite{wang2022contrastive},\cite{herreragerena2022claws}, a widely employed technique in the realm of Self-Supervised Learning \cite{kumar2023discerning}. What sets our approach apart is its deliberate departure from dependence on pre-trained feature encoders, data label annotations, or supervised pre-trained backbones. This distinctive feature underscores the autonomous and self-sufficient nature of our model, positioning it as a pioneering method in the landscape of unsupervised image retrieval. By eschewing reliance on labeled data and pre-training, our model exhibits a remarkable level of self-sufficiency, offering a novel perspective in the exploration of unsupervised learning paradigms.

\textbf{Self-supervised Learning:} Recent strides in self-supervised and unsupervised representation learning have been marked by the ascendancy of contrastive learning, an influential paradigm in this domain. Prior works [\cite{chen2020simple}, \cite{he2020momentum}, \cite{wei2020co2}, \cite{8578491}] have extensively explored this approach, wherein robust random augmentation is applied to each input image to generate positive counterparts. Following this, a contrastive loss function is employed to bring positive counterparts closer while simultaneously separating them from negative counterparts, where distinct instances serve as negatives.Preceding the adoption of contrastive learning, a prevalent methodology involves the formulation of various pretext tasks. These tasks, ranging from predicting image rotations \cite{gidaris2018unsupervised} to solving jigsaw puzzles \cite{noroozi2017unsupervised}, are designed to create self-supervision signals. These signals play a crucial role in facilitating unsupervised representation learning by providing the necessary auxiliary information for the model.
Our approach leverages Contrastive learning in its most unadulterated form, focusing on the identification of similarities. It employs a nuanced strategy by designating the cropped image as the query image for the purpose of image retrieval. This meticulous utilization of Contrastive learning aims to discern and emphasize image similarities through a thoughtful integration with the cropped image, thereby optimizing the efficiency of the image retrieval process.

\textbf{Visiolinguistic Pre-training Approach:} An intriguing realm of inquiry revolves around the task of composed image retrieval, where the search query comprises an image-language pair \cite{liu2021image}, \cite{vo2018composing} \cite{9093421} \cite{9157634}. This approach distinguishes itself by incorporating vision and language pretraining to enhance the intricacies of the retrieval process. What sets this methodology apart is its departure from the traditional paradigm of training all-encompassing models on task-specific datasets from the ground up. Instead, it embarks on its journey with representations extracted from a considerably expansive image-text corpus, often complemented by subsequent fine-tuning for task-specific objectives.

An exemplar in this landscape is the CLIP model \cite{radford2021learning}, which adopts a visiolinguistic approach to augment its retrieval capabilities. This model serves as a testament to the potential of synergizing vision and language, demonstrating the capacity to cultivate more robust and versatile image retrieval systems. The departure from conventional training methods and the integration of vision and language pretraining open avenues for heightened adaptability and effectiveness in addressing the nuances of composed image retrieval tasks.

\subsection{Contributions}
\label{contributions}
To summarize, we have the following contributions to this work:
\begin{itemize}
    \item We have formulated a novel algorithm for object retrieval that employs a fully unsupervised learning approach to "Learn" embeddings and subsequently "Search" analogous objects.
    \item We present a comprehensive comparison of our method against several augmentations and evaluate the effectiveness of our methods on various parameters.
\end{itemize}

\section{Our Method}
\label{method}



\begin{figure}[t]
  \centering
   \includegraphics[width=\linewidth]{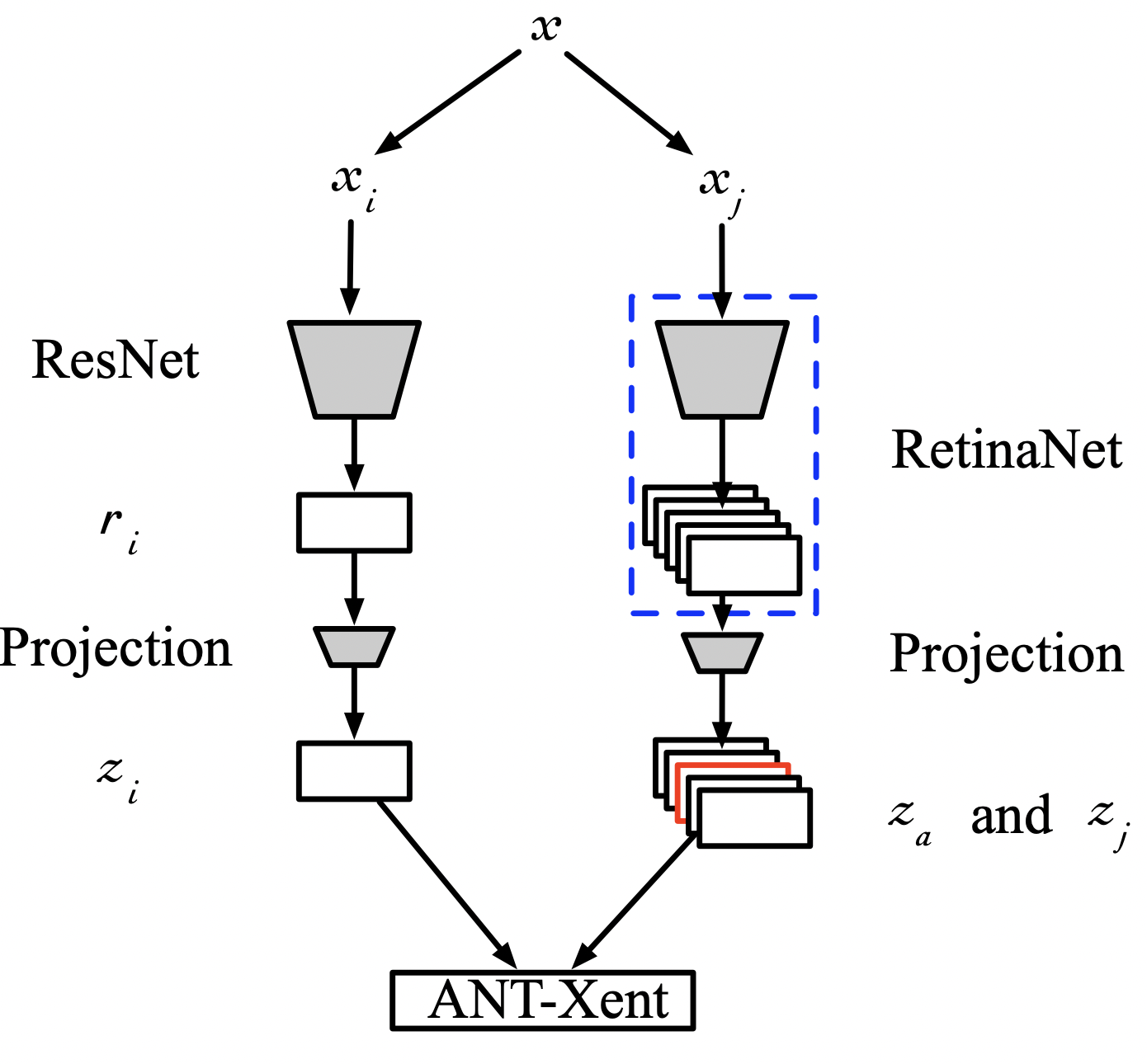}

   \caption{Flowchart for our methodology. $x$ is input image; $x_i$ is the random crop of original image($x$) and $x_j$ is the augmented image; $r_i$ is the representation from resnet; $z_i$ and $z_j$ and are the embeddings from $x_i$ and $x_j$ respectively; $z_a$ is the anchor embedding}
   \label{fig:flowchart}
\end{figure}

Efficiently distinguishing between images necessitates a feature vector or image representation endowed with discriminative qualities. This representation must not only encapsulate discriminative features but also exhibit resilience to specific transformations. These dual attributes lay the foundation for a robust similarity measure between two images, ensuring an accurate reflection of their semantic relevance.

In the overarching landscape of self-supervised learning (SSL), our primary objective is to distill generalized features from extensive volumes of unlabeled data, emancipating the learning process from the confines of specific tasks. The acquired features subsequently empower the execution of diverse downstream tasks, requiring minimal supervised training and only a limited set of task-specific labeled data.

To address the SSL objective, contrastive learning emerges as a prevalent and effective strategy. This approach revolves around learning features that withstand the impact of data augmentations applied to the input data. The rationale behind this lies in the understanding that data augmentations predominantly pertain to the style space, often bearing negligible consequences for downstream tasks. In our specific implementation, we adopt a contrastive learning pipeline, following the approach elucidated in \cite{kumar2024unsupervised}. This pipeline integrates ResNet \cite{he2015deep} into one branch and RetinaNet \cite{lin2018focal} into the other, leveraging the strengths of both for a comprehensive exploration of the image space. We extend the algorithm used in \cite{kumar2024unsupervised}  to add projection head to Pipeline 2 (Retinanet) to enhance the learning process.

The figure presented as \autoref{fig:flowchart} illustrates the stepwise flowchart for our methodology, providing a visual representation of the process. The diagram delineates the dynamic interaction between our two pipelines. In the left pipeline, designated as Pipeline 1, the process commences with input data $x_i$, undergoes image processing, and culminates in the creation of a representation tailored for integration into our ANT-Xent loss. Simultaneously, the right pipeline, denoted as Pipeline 2, operates on input $x_j$, undergoing a series of image processing operations that conclude with the extraction of Feature Pyramid Network (FPN) outputs. These FPN outputs are meticulously curated to discern positive and negative samples within the image, as illustrated below.

In the realm of losses applied to facilitate discriminative learning for different feature learning approaches, various innovative strategies have been employed. Siamese-based loss functions \cite{g2016learning} aim to minimize the global loss, leading to discriminative feature learning. Additionally, triplet quantization loss \cite{zhou2017deep}, applied in deep hashing, seeks to find the similarity between anchor positive pairs and anchor negative pairs. The NT-Xent loss function \cite{chen2020simple} introduces a temperature scaling parameter to either smooth out or accentuate the output. However, none of these loss functions incorporates a location parameter.

For the integration of location information, we introduce the Anchor-based NT-Xent Loss function (\textit{ANT-Xent}) \cite{kumar2024unsupervised}. This loss function incorporates anchor negatives in addition to anchor positives and anchor negatives generated by the NT-Xent loss. The equation for the Anchor Negative ($\textit{AN}$) is defined as follows:

\begin{equation}
\label{eq:AN}
\begin{aligned}
\textbf{\textit{AN}} &= \sum_{k=1}^{A} \exp(\text{sim}(z_i, z_k)/\tau)
\end{aligned}
\end{equation}

Here, $\textbf{\textit{AN}}$ represents Anchor Negatives, $A$ denotes the set of new negative anchors, and $\tau$ signifies the temperature parameter.

In Equation~\eqref{eq:AN}, we see the definition of $\textbf{\textit{AN}}$, allowing the pipeline to perform location-based contrast for each crop. These anchor negatives are generated in addition to the anchor positives and anchor negatives generated by the NT-Xent loss.

The complete loss function is defined as:

\begin{equation}
\begin{aligned}
\textit{$l_{i,j}$} &= -\log\frac{\exp(\text{sim}(z_i, z_j)/\tau)}{\sum_{k=1}^{2N} 1_{k\neq i} \exp(\text{sim}(z_i, z_k)/\tau) + \textit{AN}}
\end{aligned}
\end{equation}

Here, $\textit{$l_{i,j}$}$ represents the loss between samples $i$ and $j$, \textit{AN} is the Anchor Negative from Equation~\eqref{eq:AN}, $N$ is the number of samples, and $\tau$ is the temperature parameter.

With this comprehensive pipeline, our primary aim is to maximize the similarity between the Image Crop generated by the ResNet pipeline and the RetinaNet \cite{lin2018focal} pipeline. The RetinaNet pipeline encompasses full-size images, and the generated crop serves as the reference object for comparisons across the entire dataset. By leveraging ResNet as the backbone, a widely adopted feature extractor in various image classification and detection pipelines, our pipeline ensures robust feature extraction. The choice of RetinaNet aligns seamlessly with our intention to conduct similarity comparisons based on the crop, given its demonstrated efficacy with dense and small-scale objects.

The contrastive learning pipeline not only facilitates unsupervised learning but also enables the search for similar images, as illustrated in the figure. This search process involves identifying images across the dataset that exhibit high similarity to the generated crop, demonstrating the versatility and utility of our methodology.

\subsection{Augmentations}
\label{augmentations}

This study involves an exhaustive exploration of diverse augmentations, mirroring the approach outlined in \cite{radford2021learning}. Our augmentation pipeline is a sophisticated amalgamation of techniques, encompassing random cropping and zooming (with a 0.65 probability), aspect ratio distortion, downsizing, and upscaling to varying resolutions, as well as minor rotations (Horizontal and Vertical Flip with a 0.5 probability), JPEG Compression (with a 0.7 probability), and HSV color jitter. Notably, the pipeline is designed with a thoughtful consideration for variability, ensuring a rich and comprehensive set of transformations.

Furthermore, our augmentation pipeline features a dynamic selection from various interpolation algorithms at each pertinent step, adding an additional layer of variability to the augmentation process. This deliberate variation in interpolation methods contributes to the robustness of our experiments.

\begin{figure*}
  \centering
  \begin{subfigure}{\linewidth}
    \includegraphics[width=1\linewidth]{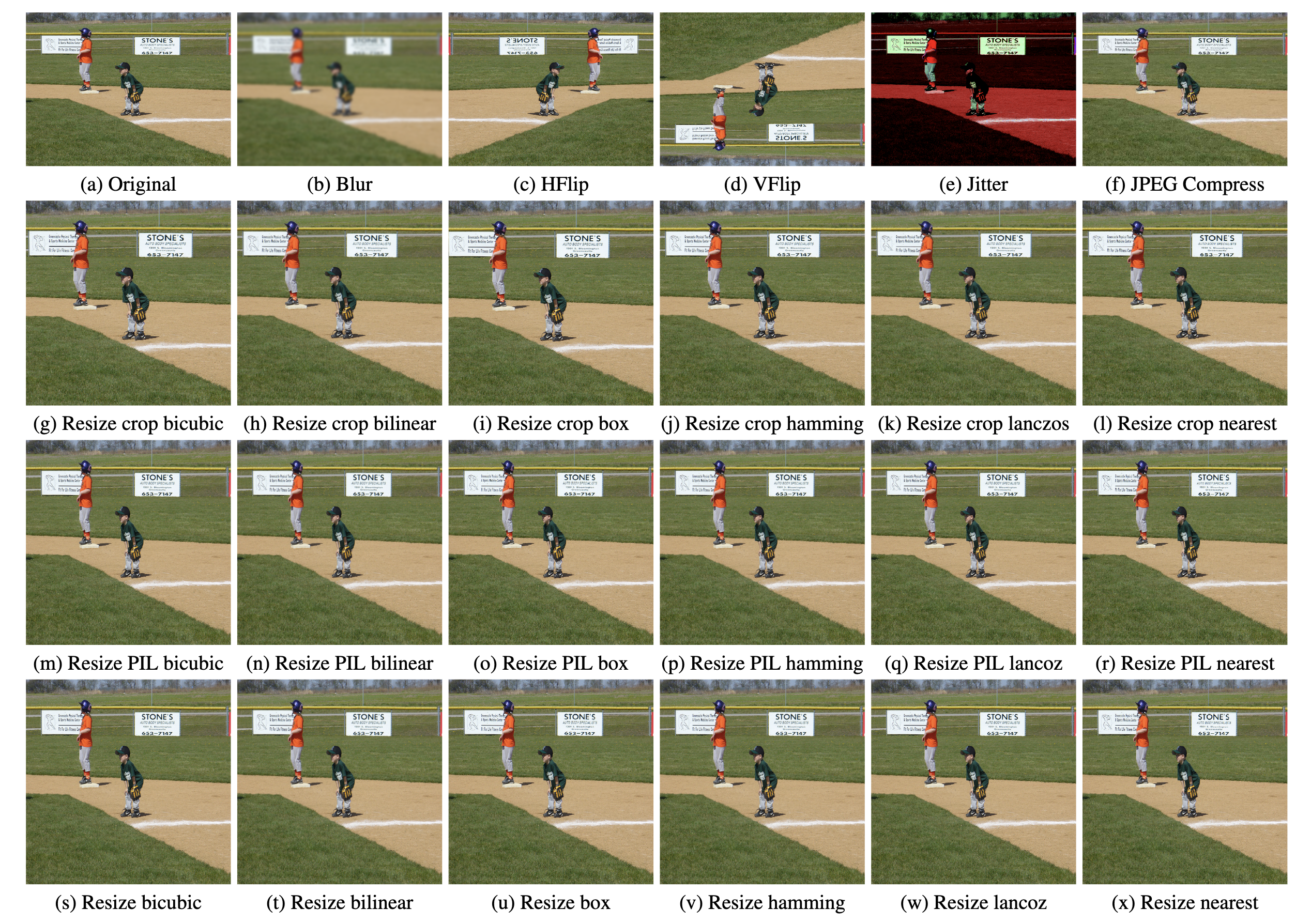}
    \caption{Original image with multiple augmentations as well as their interpolations}
    \label{fig:short-a}
  \end{subfigure}
  \hfill
  \begin{subfigure}{0.5\linewidth}
    \includegraphics[width=\linewidth]{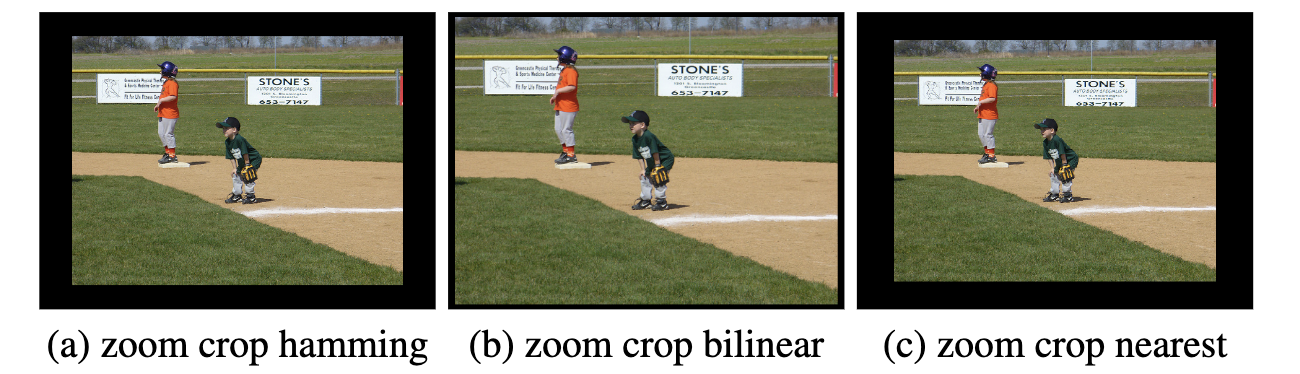}
    \caption{Zoom crop augmentation with interpolation}
    \label{fig:short-b}
  \end{subfigure}
  \caption{We have used augmentations extensively in our experiments. In this visualization, we present all the augmentations we have used in the experiments along with all the interpolations.}
  \label{fig:short}
\end{figure*}

For a visual representation of the augmentations employed in our experiments, refer to \autoref{fig:short}, which comprises both \autoref{fig:short-a} and \autoref{fig:short-b}. This comprehensive visual depiction serves as a detailed reference, encapsulating the spectrum of augmentations and interpolations utilized throughout our experimentation process.





\section{Experiments}
\label{experiment}



In our experimental framework, we have established four distinct models, each strategically designed to bolster the learning process. The fundamental goal underlying these models is to refine and augment the learning experience. In the first model (Model 1), an array of augmentations is employed, with a particular emphasis on color jitter. Notably, we introduce an element of randomness in the initialization of color jitter parameters, aiming to meticulously scrutinize and understand the consequential impact on the learning process.

Moving on to Model 2, we adopt a more controlled approach. Here, the color jitter parameters are deliberately fixed to optimal values, facilitating a focused investigation into their influence on the overall results. This deliberate fixation allows us to isolate and comprehend the specific contribution of color jitter within the learning dynamics.

Model 3 delves into the manipulation of Gaussian Blur, enhancing its potency within the augmentation process. Furthermore, we introduce additional complexity by incorporating crops with random interpolation. This model is engineered to explore the nuanced interplay between Gaussian Blur, random interpolation, and their cumulative effect on the learning outcomes.

In the final model (Model 4), we introduce a pivotal component—projection heads. The integration of projection heads serves the purpose of refining the model's representations by projecting backbone features into a low-dimensional space before the application of the loss function. Extensive experiments conducted by \cite{gupta2022understanding} underscore the consistent enhancement in performance achieved through the incorporation of projection heads. This observation reinforces the utility of projection heads in refining and optimizing feature representations.
We train all our models for 100 epochs except \textit{Model 1} on which we performed early stopping and limited to 80 epochs.

It is pertinent to note that all our experiments are conducted exclusively on images sourced from MS-COCO \cite{lin2015microsoft}. The evaluation process is meticulously executed in a zero-shot manner, where no fine-tuning of labels or data is undertaken. This deliberate approach ensures a stringent examination of the models' capabilities in discerning regions of utmost similarity within an image relative to a cropped image. The zero-shot evaluation methodology underscores the robustness and adaptability of our models, emphasizing their potential in real-world scenarios where fine-tuning resources may be limited or impractical.

\section{Results}
\label{results}
\subsection{Accuracy for "Learn"}
\label{learn_accuracy}
Our extensive experiments, incorporating various augmentations, form the basis for a comprehensive analysis of their effectiveness in facilitating unsupervised learning scenarios. These augmentations were meticulously chosen and tested to discern their impact on learning dynamics. The subsequent discussion delves into the results obtained, with a specific focus on the layer-wise Similarity Grid Accuracy (SGA) across different models as presented in Table \ref{sample-table}
\begin{table*}[t]
\centering 
\begin{tabular}{|c|c|c|c|c|c|c|} 
\hline
\bf \textit{Model} &\bf \textit{Method} & \bf \textit{Layer 0} & \bf \textit{Layer 1} & \bf \textit{Layer 2} & \bf \textit{Layer 3} & \bf \textit{Layer 4}\\
\hline
Model 1 & Color Jitter + JPEG Compresssion (random) & 0.3136 & 0.3132 & 0.3092 & 0.3136 & 0.2944 \\
Model 2 & Color Jitter + JPEG Compression(optimised) & 0.846 & 0.8456 & 0.8484 & 0.7952 & 0.66 \\
Model 3 & Gaussian Blur + Crop with Random interpolation & 0.8324 & 0.8356 & 0.8324 & 0.7984 & 0.6948 \\
Model 4 & Projection Head & 0.8388 & 0.8432 & 0.8344 & 0.7984 & 0.6908 \\
\hline
\end{tabular}
\caption{Similarity Grid Accuracy (Higher is better) per layer of FPN }
\label{sample-table}
\end{table*}


Table \ref{sample-table} offers a detailed breakdown of the Similarity Grid Accuracy (SGA) for each layer of the Feature Pyramid Network (FPN) across various models. The FPN serves as a crucial backbone for the RetinaNet model, employed in pipeline 2 of our experimental workflow.

The provided SGA values are indicative of the models' proficiency in discerning similarities within an image concerning a cropped section. Each row corresponds to a different model, and each column represents a specific FPN layer (Layer 0 to Layer 4). The SGA values, representing the accuracy of similarity grid predictions, are detailed for each layer within the FPN.

    \begin{itemize}
        \item Model 1: This model employs Color Jitter and random JPEG Compression. The SGA values across layers range from 0.2944 to 0.3136.
        \item Model 2: Utilizing Color Jitter and optimized JPEG Compression, Model 2 showcases significantly higher SGA values, ranging from 0.66 to 0.8484.
        \item Model 3: Incorporating Gaussian Blur and crops with random interpolation, Model 3 achieves SGA values spanning from 0.6948 to 0.8356.
        \item Model 4: With the inclusion of a Projection Head, Model 4 consistently outperforms other models, exhibiting SGA values ranging from 0.6908 to 0.8432 across different layers.
    \end{itemize}

The SGA values serve as a quantitative measure of how well each model, at different layers of the FPN, excels in recognizing similarities between images and their corresponding cropped sections. Notably, Model 4 with the Projection Head consistently demonstrates superior performance across nearly all layers compared to the other models, suggesting the efficacy of the Projection Head in enhancing feature representations within the FPN. This observation aligns with the findings in \cite{kumar2024unsupervised}, reinforcing the value of the Projection Head in optimizing the model's ability to capture and understand visual similarities.

\begin{figure*}
  \centering
  \begin{subfigure}{0.33\linewidth}
    \includegraphics[width=\linewidth]{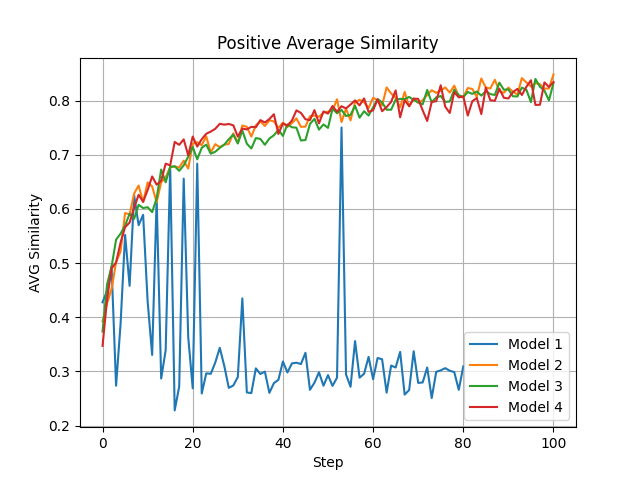}
    \caption{Average Positive}
    \label{fig:avg_pos}
  \end{subfigure}
  \hfill
  \begin{subfigure}{0.33\linewidth}
    \includegraphics[width=\linewidth]{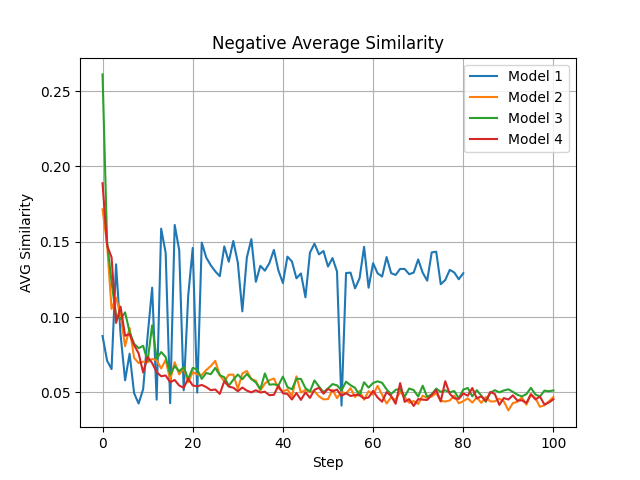}
    \caption{Average Negative}
    \label{fig:avg_neg}
  \end{subfigure}
  \hfill
  \begin{subfigure}{0.33\linewidth}
    \includegraphics[width=\linewidth]{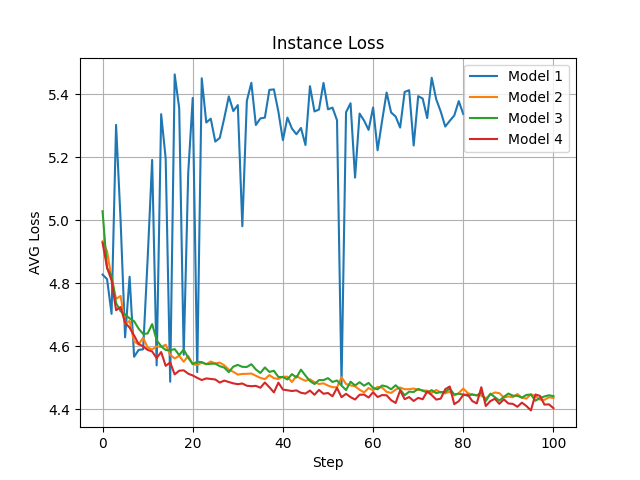}
    \caption{Average Loss}
    \label{fig:loss}
  \end{subfigure}
  \caption{Performance of different models}
  \label{fig:graphs}
\end{figure*}



In \autoref{fig:graphs}, we present an extensive comparative analysis of the similarity behavior exhibited by each model. To delve into the intricacies, \autoref{fig:avg_pos} meticulously illustrates the behavior concerning representations produced from the same image—essentially capturing the similarity between positive pairs. The nuances of the model's discernment become apparent in \autoref{fig:avg_neg}, where the comparison extends to representations from different image pairs. In this scenario, low values are anticipated, signifying a desired outcome where the similarity between image $x_i$ and $x_j$ is minimal.

To further enrich our understanding, \autoref{fig:loss} provides a visual representation of the loss incurred by each of the models during the training phase. These plots are derived from the evaluation dataset, meticulously gathering specific metric values at each step and subsequently averaging them at the conclusion of each epoch. This comprehensive exploration of similarity metrics and loss dynamics offers a detailed insight into the performance nuances exhibited by the different models under scrutiny.

\subsection{Accuracy for "Search"}
\label{search_accuracy}
\begin{table}
  \centering
  \begin{tabular}{@{}l@{\hspace{20pt}}c@{}@{\hspace{20pt}}c@{}@{\hspace{20pt}}c@{\hspace{10pt}}}
    \toprule
    \bf \textit{Model} & \bf \textit{Top 1} & \bf \textit{Top 5} & \bf \textit{Top 10} \\
    \midrule
    Model 1 & 0.04 & 0.20 & 0.24 \\
    Model 2 & 0.26 & 0.57 & 0.71 \\
    Model 3 & 0.18 & 0.41 & 0.54 \\
    Model 4 & 0.27 & 0.52 & 0.63 \\
    \bottomrule
  \end{tabular}
  \caption{Top Accuracy for every model used in the experiment}
  \label{tab:accuracy}
\end{table}

In addition to evaluating the models' performance through layer-wise Similarity Grid Accuracy (SGA), we delve into their classification capabilities with a focus on Top-1, Top-5, and Top-10 accuracy metrics. Table \ref{tab:accuracy} provides a comprehensive comparison across all four models. This multi-faceted analysis enables a holistic understanding of the models' proficiency not only in discerning similarities but also in accurate image classification. The ensuing results section dissects these findings, shedding light on the nuanced strengths and capabilities exhibited by each model in both unsupervised learning scenarios and classification tasks.

\begin{figure*}[t]
  \centering
   \includegraphics[width=\linewidth]{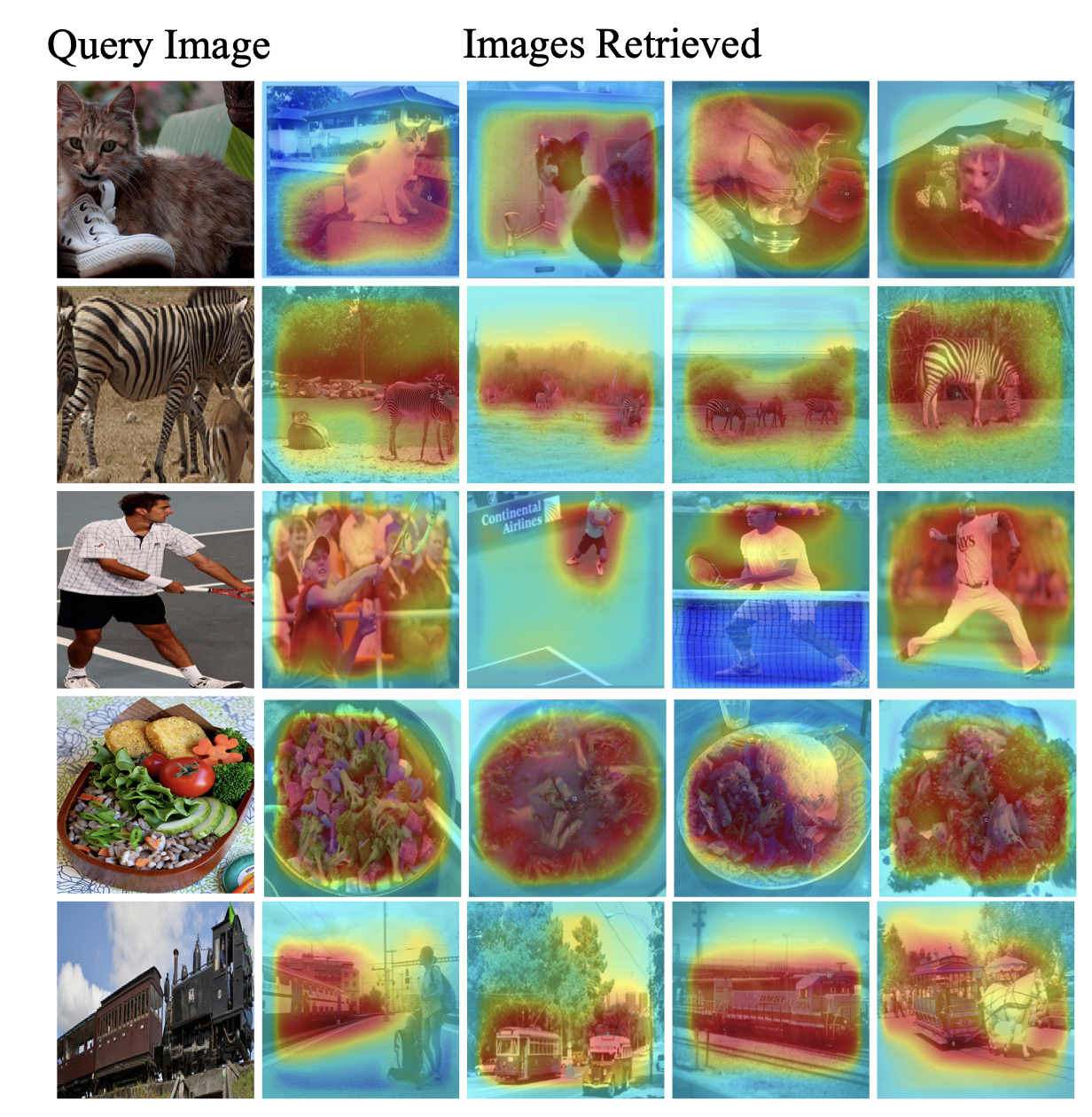}

   \caption{This figure shows a grid of images gathered after selecting a crop within the dataset and searching the top10 similar images. The selected crop is passed to the RetinaNet to produce a representation and the highest similarity images are process on a batch to produce the FPN outputs used to compare and execute the selection.}
   \label{fig:topk}
\end{figure*}

The visual representation depicted in \autoref{fig:topk} intricately portrays the accuracy of our "Search" mechanism. Notably, the figure showcases the intricate process wherein the system conducts searches to identify the top 10 images exhibiting the highest similarity with the original image. This original image is derived from a random crop, denoted as $x_i$, within our comprehensive methodology, as illustrated in the detailed flowchart presented in \autoref{fig:flowchart}. The color gradient observed in the retrieved images serves as a visual indicator, where the hue transitions from red, denoting the highest similarity to the input image, to blue, indicative of the least similarity. This gradient offers a nuanced and illustrative representation, effectively conveying the varying degrees of resemblance between the retrieved images and the input image.

\section{Conclusion}
\label{conclusion}

In the course of this investigation, we introduce "Learn and Search," an innovative methodology meticulously crafted to enhance the efficiency and efficacy of object retrieval systems through cutting-edge advancements in object recognition and segmentation.

The escalating deluge of digital content underscores the imperative for inventive solutions capable of facilitating robust object lookup and retrieval. "Learn and Search" emerges as a watershed in this landscape, harnessing the potency of contrastive learning to address the intricate challenges entwined with object search. By seamlessly integrating deep learning principles and contrastive learning techniques, our approach not only signifies a paradigm shift but also charts a transformative trajectory within domains such as image recognition, recommendation systems, and content tagging.

The symbiosis of deep learning and contrastive learning heralds a monumental stride in content-based search and retrieval. "Learn and Search" not only augments the precision of object retrieval systems but also paves the way for a more refined era in digital content management. As technology perpetually evolves, our endeavor stands as a compelling illustration of the ongoing refinement and evolution of methodologies within the dynamic sphere of object classification and segmentation. The presented approach, poised at the nexus of innovation and practicality, is poised to contribute substantively to the continuous progression of inventive solutions in the ever-expanding arena of digital content analysis and retrieval.

{
    \small
    \bibliographystyle{ieeenat_fullname}
    \bibliography{main}
}


\end{document}